\documentclass{article}
\usepackage{spconf,amsmath,graphicx,array,xcolor,bm,url,booktabs}
\usepackage[ruled,vlined]{algorithm2e}

\definecolor{myblue}{rgb}{0,0.5,1}

\title{End-to-End Attention based Text-Dependent Speaker Verification}
  \name{Shi-Xiong Zhang, Zhuo Chen$^{\dag}$, Yong Zhao, Jinyu Li and Yifan Gong}
  \address{Microsoft Corporation, One Microsoft Way, Redmond, WA 98052 \\
  $^{\dag}$Columbia University, New York, NY, USA \\
  {\small \tt \{zhashi, yonzhao, jinyli, ygong\}@microsoft.com, }{$^{\dag}$\small \tt zc2204@columbia.edu}}

\begin{document}
%\ninept
\sloppy
\maketitle
\begin{abstract}
A new type of End-to-End system for text-dependent speaker verification is presented in this paper. Previously, using the phonetic/speaker discriminative DNNs as feature extractors for speaker verification has shown promising results. The extracted frame-level (DNN bottleneck, posterior or d-vector) features are equally weighted and aggregated to compute an utterance-level speaker representation (d-vector or i-vector). In this work we use speaker discriminative CNNs to extract the noise-robust frame-level features. These features are then combined to form an utterance-level speaker vector through an attention mechanism. The proposed attention model takes the speaker discriminative information and the phonetic information to learn the weights. The whole system, including the CNN and attention model, is joint optimized using an end-to-end criterion. The training algorithm imitates exactly the evaluation process --- directly mapping a test utterance and a few target speaker utterances into a single verification score. The algorithm can automatically select the most similar impostor for each target speaker to train the network. We demonstrated the effectiveness of the proposed end-to-end system on Windows $10$ ``Hey Cortana" speaker verification task.
\end{abstract}

\section{Introduction}
\label{sec:intro}

\begin{figure}[bh]
        \centering
        \includegraphics[width=1.0\linewidth]{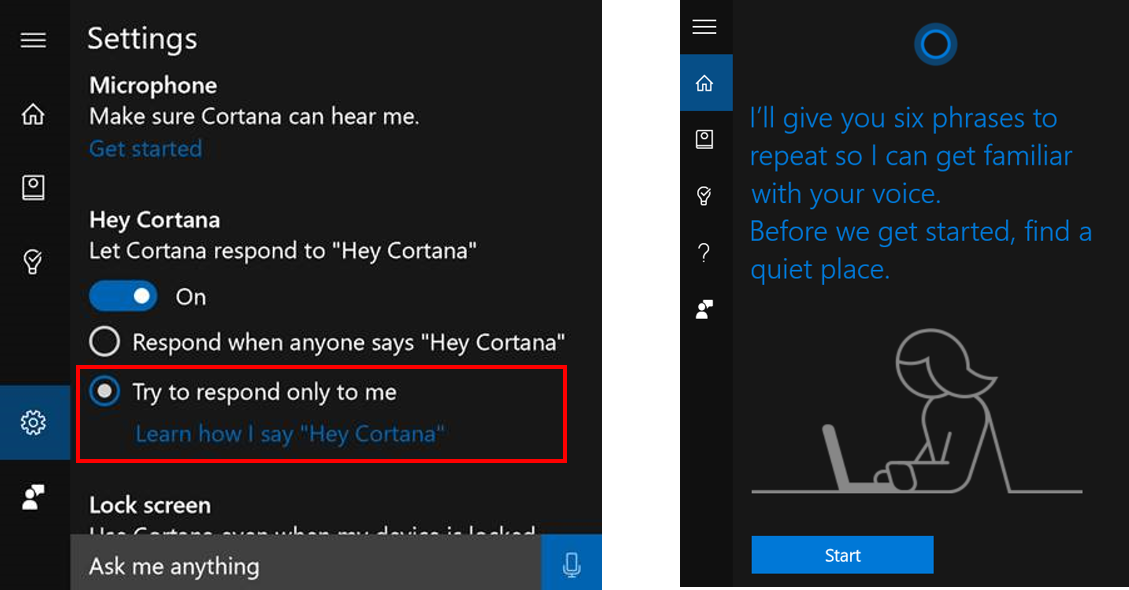}
        \caption{{\it The keyword spotting and speaker verification (with the fixed phase ``Hey Cortana") in Windows 10.}}
        \label{fig:HC}
\end{figure}

Speaker verification (SV) is a binary classification problem in which a person's identity is verified based on his/her voice. Speaker verification can be categorized into \emph{text-dependent} and \emph{text-independent} \cite{Campbell97}.
In text-dependent systems, the same set of phrases are used for enrollment and recognition. In text-independent systems, on the other hand, different phrases are used. The text-dependent SVs usually outperforms the text-independent SVs, because of the constraint of the phonetic variability \cite{Auckenthaler99,Matsui&Furui91,RSR2015}. 
Especially with the increasing popularity of mobile/wearable devices, it is competitively beneficial to enable the full voice interaction beginning from a fixed-phrase (keyword) voice-authenticated wake-up \cite{End2EndGoogle}. At Microsoft, we are interested in the \emph{text-dependent} speaker verification with the global keyword ``Hey Cortana" (see Fig. \ref{fig:HC}). 
Mathematically, the relationship between text-dependent/independent SV and Keyword Spotting (KWS) wake-up system can be described in the following equation
%precise and reliable alignment between the unknown speech and reference templates can be made.
\begin{align}
\overbrace{P(\textcolor{red}{\sf{spk}}|{\mathbf O}_{\sf 1:T})}^{\text{text-independent SV}}&= \sum\limits_{\textcolor{blue}{{\bf w}}} \overbrace{P(\textcolor{red}{\sf spk}, \textcolor{blue}{{\bf w}_{\sf 1:L}} | {\mathbf O}_{\sf 1:T})}^{\text{joint speaker \& KWS wakeup}} \nonumber\\
                         &= \sum\limits_{\textcolor{blue}{{\bf w}}} ~\underbrace{P(\textcolor{red}{\sf spk} | \textcolor{blue}{{\bf w}_{\sf 1:L}}, {\mathbf O}_{\sf 1:T})}_{\text{text-dependent SV}} 
                         \underbrace{P(\textcolor{blue}{{\bf w}_{\sf 1:L}}| {\mathbf O}_{\sf 1:T})}_{\text{KWS speech recognition}} \nonumber
\end{align}
where $\mathbf O_{\sf 1:T}$ is the observation sequence and $\textcolor{blue}{{\bf w}_{\sf 1:L}}$ is stand for the word/phone/state sequence.  One major advantage of \emph{text-dependent} speaker verification systems is it has knowledge of the utterance’s phonetic content and can achieve robust verification results with very short enrollment utterances \cite{variani2014deep}.

Previously, the traditional techniques \cite{Reynolds95b,Kenny07} used for \emph{text-independent} speaker verification have been found ineffective for the \emph{text-dependent} tasks \cite{RSR2015}. Better performance can be achieved by slightly modifying the older techniques such as GMM-UBM \cite{RSR2015}, GMM-SVM \cite{novoselov2014text,zhang2011optimized} and  i-vector/PLDA \cite{stafylakis2013text}.
More recently, the deep neural networks (DNNs) \cite{lei2014novel} and recurrent neural networks (RNNs) \cite{Bhattacharya+2016} have been successfully applied to \emph{text-dependent} speaker verification. Two types of DNNs, speaker discriminative DNNs \cite{variani2014deep} and phonetic discriminative DNNs \cite{richardson2015deep}, have been investigated to extract the frame-level features, such as d-vectors \cite{variani2014deep}, bottleneck features \cite{richardson2015deep,Zeinali+2016}
and phonetic posterior features (alignments) \cite{lei2014novel,Zeinali+2016}. These features are treated equally important and then aggregated together to compute an utterance-level speaker representation, such as i-vectors \cite{lei2014novel,Zeinali+2016},  d-vectors \cite{End2EndGoogle,Bhattacharya+2016}. These utterance-level features from the test speaker and enrolled speaker are then scored using a similarity measure, e.g., cosine distance/PLDA \cite{dehak2010cosine,kenny2013plda}. The DNNs/RNNs used to extract phonetic/speaker features are fixed after the training stage.  Google recently proposed an End-to-End method to train the DNNs/RNNs for text-dependent speaker verification \cite{End2EndGoogle}. This is in contrast to the established approach of training DNNs to discriminate between speakers at the frame-level. In \cite{End2EndGoogle} the cosine distance score of two utterance-level representations are passed to a logistic regression layer to produce the final loss $-\log P(\sf{accept/reject})$. The parameters of whole networks are learned by minimizing this end-to-end loss.

In this paper we use speaker discriminative CNNs to extract noise robust frame-level features. This is inspired by recent works in \cite{sercu2016advances,saon2015ibm} which illustrated a deep convolution neural network (CNN) architecture outperformed LSTMs in many speech recognition tasks. Another contribution of this work is the CNN extracted frame-level features are smartly combined through an attention mechanism to generate an utterance-level speaker representation, instead of just equally weighting and averaging all the frames \cite{End2EndGoogle,variani2014deep}. The proposed attention model takes the speaker discriminative information and phonetic information to learn the attention weights. The third contribution is the whole system, including the CNN and attention model, is joint optimized using a novel end-to-end training algorithm. Unlike the Google's end-to-end training \cite{End2EndGoogle} which randomly samples the test speaker and the target speaker, our algorithm uses the most competing impostor for each target speaker (in the case of rejection). Finally, the end-to-end system is evaluated on Windows 10 ``Hey Cortana" speaker verification task.

\section{End-to-End Speaker Verification}
\label{sec:end2end}

This section describes the overview architecture of our end-to-end speaker verification system. The detail of its important components will be discussed in Section \ref{sec:DNN} and \ref{sec:attention}. The experimental results and analysis can be found in Section \ref{sec:Exp}.

\subsection{End-to-End architecture}
\label{ssec:arch}

\begin{figure}[t]
%\begin{minipage}[b]{1.0\linewidth}
  %\centering
  \hspace{0.2cm}
  \centerline{\includegraphics[width=9.5cm]{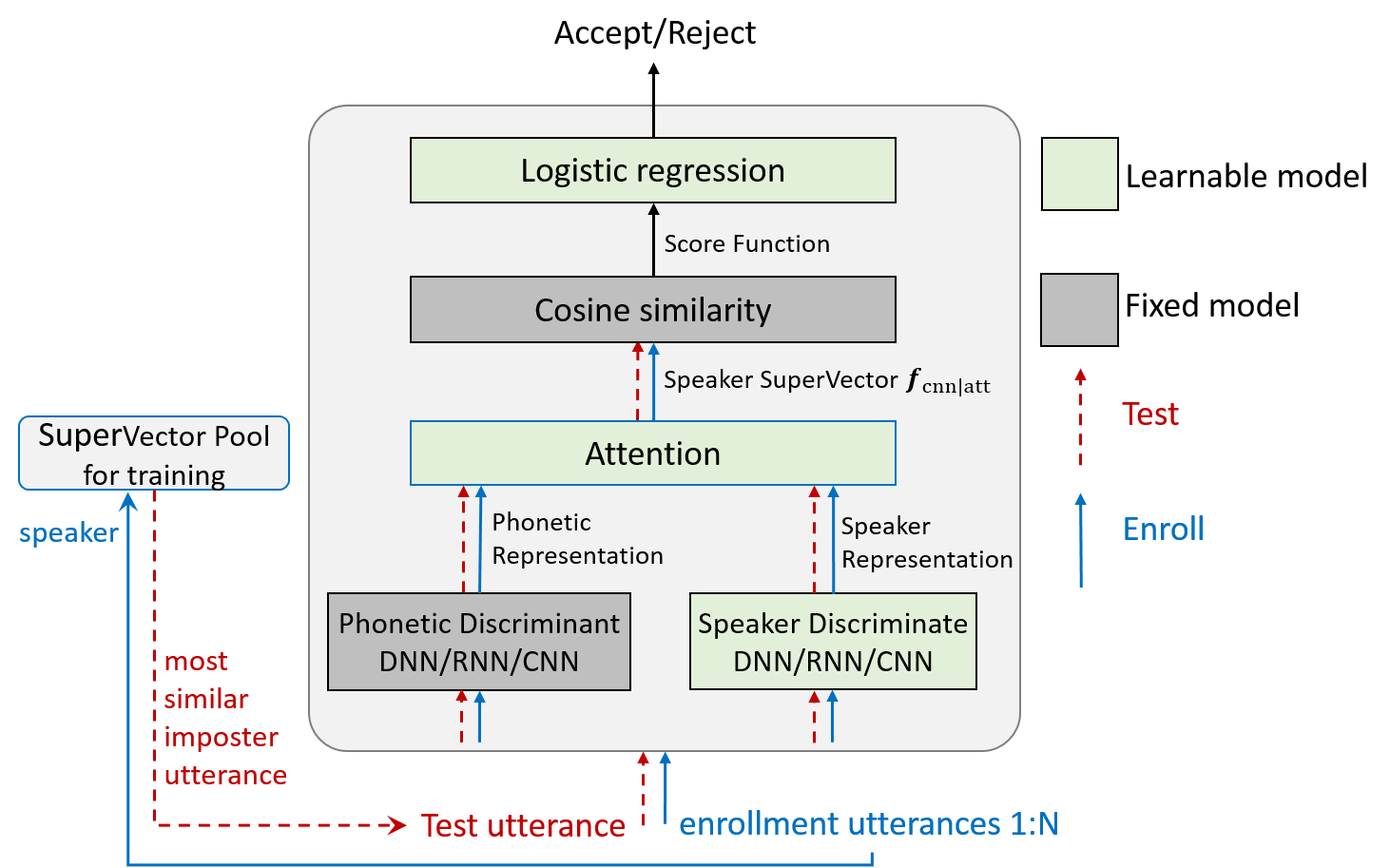}}
%  \vspace{2.0cm}
%  \centerline{(a) Result 1}\medskip
%\end{minipage}
% %
% \begin{minipage}[b]{.48\linewidth}
%   \centering
%   \centerline{\includegraphics[width=4.0cm]{image3}}
% %  \vspace{1.5cm}
%   \centerline{(b) Results 3}\medskip
% \end{minipage}
% \hfill
% \begin{minipage}[b]{0.48\linewidth}
%   \centering
%   \centerline{\includegraphics[width=4.0cm]{image4}}
% %  \vspace{1.5cm}
%   \centerline{(c) Result 4}\medskip
% \end{minipage}
% %
\caption{The architecture of our end-to-end attention based system for speaker verification. The green blocks indicate the models can be learned in the training phase. All the green and grey blocks are fixed in the enrollment and verification phases. During enrollment, the extracted speaker supervectors are stored as a speaker model. The red dash and blue solid arrows indicate the representations of the test utterance and enrolled utterances, respectively. The speaker vector pool contains all the speaker representations in the training set and is used to select the most similar impostor during training. 
}
\label{fig:end2end}
\end{figure}

A typical speaker verification protocol includes three phases: training, enrollment, and evaluation. 
In the training phase, our network learns to extract an internal speaker representation from the utterance. The learning network includes two parts, a CNN and an attention network, as shown in Figure \ref{fig:end2end}. This network learning stage is like the UBM training stage. The details of our CNN model and attention network will be discussed in Section \ref{sec:DNN} and \ref{sec:attention}. Note that the networks are not trained to discriminate between speakers at the frame-level. Instead, all the parameters in the whole system will be jointly trained using an end-to-end criterion described in Section \ref{ssec:train}.

After training, the CNN and attention networks are used to extract the utterance-level speaker vectors (see Fig. \ref{fig:end2end}) for the enrolment and test speakers. This is done by freezing the parameters of the learning network. In the enrollment phase, each speaker provides a few utterances and each utterance is converted to a supervector through the trained network. The speaker model is obtained by averaging over a small mount of enrollment suptervectors.

% Below is an example of how to insert images. Delete the ``\vspace'' line,
% uncomment the preceding line ``\centerline...'' and replace ``imageX.ps''
% with a suitable PostScript file name.
% -------------------------------------------------------------------------

In the evaluation phase, a scoring function $S({\bf O}_{\textcolor{red}{\sf tst}}, {\bf O}^{\sf 1:N}_{\textcolor{blue}{\sf spk}})$ is used to compute the similarity between a test utterance $\bf O_{\textcolor{red}{\sf tst}}$ and a claimed speaker $\sf spk$. The final score will be compared against a pre-defined (may be speaker dependent) threshold. The system will accept if the score exceeds the threshold, i.e., the utterance $\bf O_{\textcolor{red}{\sf tst}}$ belongs to $\sf spk$, and reject otherwise. 
Typically, a simple scoring function is the cosine similarity \cite{dehak2010cosine} between the test speaker representation ${\bm f}_{\sf cnn|att}({\bf O}_{\textcolor{red}{\sf tst}})$ and the speaker model ${\bm f}_{\sf cnn|att}({\bf O}^{\sf 1:N}_{\textcolor{blue}{\sf spk}})$,

\begin{equation}
S({\bf O}_{\textcolor{red}{\sf tst}}, {\bf O}^{\sf 1:N}_{\textcolor{blue}{\sf spk}}) = \frac{{\bm f}_{\sf cnn|att}({\bf O}_{\textcolor{red}{\sf tst}})^{\sf T} {\bm f}_{\sf cnn|att}({\bf O}^{\sf 1:N}_{\textcolor{blue}{\sf spk}}) }{ \| {\bm f}_{\sf cnn|att}({\bf O}_{\textcolor{red}{\sf tst}})\| ~\|  {\bm f}_{\sf cnn|att}({\bf O}^{\sf 1:N}_{\textcolor{blue}{\sf spk}}) \|}
\end{equation}
where the speaker model ${\bm f}_{\sf cnn|att}({\bf O}^{\sf 1:N}_{\textcolor{blue}{\sf spk}})$ is precomputed in the enrollment stage and ${\bm f}_{\sf cnn|att}(\cdot)$ indicates that the feature mapping depends on the CNN and attention networks. In \cite{End2EndGoogle} the similarity score $S$ is then passed to a logistic regression $\sigma(\cdot)$ which includes a linear layer with a bias. Thus,
\begin{equation}
     P ({\sf accept}|{\bf O}_{\textcolor{red}{\sf tst}}, {\bf O}^{\sf 1:N}_{\textcolor{blue}{\sf spk}}) 
     = \sigma\left( S \right)  
     =  \frac{1}{1+ e^{-w S({\bf O}_{\textcolor{red}{\sf tst}}, {\bf O}^{\sf 1:N}_{\textcolor{blue}{\sf spk}})-b} } 
\end{equation}
where scalar $w$ is a score normalizer trained using the criterion described in the next section, %Thus no further score normalization is needed. 
$-b/w$ can be viewed as a verification threshold. The reject probablity can be computed by $P ({\sf reject}|{\bf O}_{\textcolor{red}{\sf tst}}, {\bf O}^{\sf 1:N}_{\textcolor{blue}{\sf spk}}) = 1- \sigma\left( S \right) $.

\subsection{End-to-End training}
\label{ssec:train}

Given the current CNN and the attention model's parameters, ${\bf W}_{\textcolor{green}{\sf cnn|att}}$, the similarity score $S({\bf O}_{\textcolor{red}{\sf tst}}, {\bf O}^{\sf 1:N}_{\textcolor{blue}{\sf spk}})$ between the test utterance and the target speaker utterances can be computed and fed to the logistic regression $\sigma(\cdot)$. All the parameters in the whole network (see architecture in Fig.~\ref{fig:end2end}) can be jointly optimized using the following end-to-end criterion \cite{End2EndGoogle}
\begin{equation}\label{eq:e2eloss}
     \mathcal{F}({\bf W}_{\textcolor{green}{\sf cnn|att}}) =  - \frac{1}{I} \sum\limits_{i=1}^{I}   y_i \log \sigma(x_i)  + (1-y_i) \log \left( 1- \sigma(x_i) \right)
\end{equation}
% \[
%      \mathcal{F}({\bf W}_{\textcolor{green}{\sf nn,att}}) =  - \frac{1}{I} \sum\limits_{\textcolor{blue}{i}=1}^{I}   y_i \log  P ({\sf accept}| \textcolor{blue}{{\text{pair}}_i})  \]
%      \[
%      + (1-y_i) \log  P ({\sf reject}| \textcolor{blue}{{\text{pair}}_i}) 
% \]
where ${\bf W}_{\textcolor{green}{\sf cnn|att}}$ represents the parameters of CNN, attention model and logistic regression  $\sigma (\cdot)$ and $\mathcal{F}(\cdot)$ is a function of ${\bf W}_{\sf cnn|att}$. 
The input $x_i=S_i({\bf O}_{\textcolor{red}{\sf tst}}, {\bf O}_{\textcolor{blue}{\sf spk}})$ is a similarity score of the $i$-th pair of $({\sf tst}, {\sf spk})$, which depends on the ${\bf W}_{\textcolor{green}{\sf cnn|att}}$. The label $y_i=1$ if the testing utterance ${\bf O}_{\textcolor{red}{\sf tst}}$ belongs to $\textcolor{blue}{\sf spk}$, otherwise $y_i=0$ . $I$ is the total number of $({\sf tst}, {\sf spk})$ pairs in the training set (which is huge). To effectively use the data, an algorithm that can smartly pick the most ``confusing" pairs is proposed.  The detail of the approach is described in Alg.~\ref{alg:train}. Note the training process with the criterion in Eq.~\ref{eq:e2eloss} is actually imitating the end-to-end evaluation metric. It directly maps a test utterance and a few target utterances to a single score for verification. The whole network is trained by minimizing the end-to-end loss.

To make sure the parameters are updated using sufficient information from diversified speakers, 
we group the speakers into mini-batches. Each mini-batch contains $64$ speakers as targets. For each target speaker, we sample $N$ utterances as enrollment and $T_1$ test utterances as ``acceptance" data. 
For each target speaker, we also select $k$ most similar speakers as impostors to make sure the network can learn to discriminate between the most challenging samples. For these $k$ impostors we randomly sample $T_2$ utterances as ``reject" data. The exact number of these hyperparameters, $N, T_1, T_2$ and $k$, are discussed in Section \ref{ssec:Exp}.
In order to efficiently search the $k$ most similar impostors, a query table is build to the store the $k$ nearest neighbor for each speaker. The table is built using the all-nearest-neighbors algorithm \cite{vaidya1989ano} with the cosine-distance metric in $\mathcal{O}(k \cdot n\log n)$ time. The speaker vector pool used to compute the table is initialized by i-vectors \cite{dehak2011front}.
The pool can be updated periodically (each full sweep) using the speaker supervectors, ${\bm f}_{\sf cnn|att}({\bf O}^{\sf 1:N}_{\textcolor{blue}{\sf spk}}) $.

\SetInd{0.5em}{0.5em}
\begin{algorithm}
\DontPrintSemicolon
\KwData{10k speakers, each speaker has 10-50 utterances.}
\KwResult{speaker discriminative CNN, attention network and logistic regression model.}
%\Begin{
initial ${\bf W}_{\textcolor{green}{\sf cnn|att}}$\;
initial speaker vector pool $\mathcal V$ $\leftarrow$ i-Vectors \;
\For{each full sweep}{
\For{each minibatch in a full sweep}{
\For{each $\sf spk$ in a minibatch}{
1) sample $N+T_1$ utterances of $\sf spk$ \;
 ~~~~$N$ enrollment: ${\bf O}^{\sf 1:N}_{{\sf spk}}$ \;
 ~~~~$T_1$ test: ${\textcolor{brown}{x_i}}=S_i({\bf O}^i_{\textcolor{red}{\sf tst}}, {\bf O}^{\sf 1:N}_{\textcolor{blue}{\sf spk}}), \textcolor{brown}{y_i=1}|_{i=1}^{T_1}$ \;
2) search $k$ most similar impostors in $\mathcal V$  \;
3) sample $T_2$ utterances belongs to these $\sf imp$ \;
~~~~$T_2$ test: $\textcolor{brown}{x_i}=S_i({\bf O}^i_{\textcolor{red}{\sf imp}}, {\bf O}^{\sf 1:N}_{\textcolor{blue}{\sf spk}}), \textcolor{brown}{y_i=0}|_{i=1}^{T_2}$ \;
 4) gather $(\textcolor{brown}{x_i}, \textcolor{brown}{y_i})$, accumulate $\nabla \mathcal{F}({\bf W}_{\sf cnn|att})$
}
update $\bf W_{\sf cnn|att} \leftarrow \bf W_{\sf cnn|att} - \eta \nabla \mathcal{F}({\bf W}_{\sf cnn|att})$ \;
}
 update $\mathcal V \leftarrow  {\bm f}_{\sf cnn|att}({\bf O}^{\sf 1:N}_{{\sf spk}}) $ $\forall ~ {\sf 
 spk}$ \; 
}
\caption{End-to-End Training for SV \label{alg:train}}
\end{algorithm}

\section{Neural Nets for Speaker Verification}
\label{sec:DNN}

This section presents a survey of DNN/RNN based methods for speaker verification. A new speaker discriminative CNN model is then proposed and applied to the end-to-end system described in previous section.
DNNs/RNNs have been successfully integrated into the speaker verification paradigms. The existing neural networks for speaker verification can be categorized into two types --- \emph{phonetic discriminative} DNNs and \emph{speaker discriminative} DNNs \cite{Bhattacharya+2016} (see Fig.~\ref{fig:end2end}). These two types of DNNs will discussed in Section \ref{ssec:asr_dnn} and \ref{ssec:spk_dnn}.

\begin{figure*}[htb]
  \centerline{\includegraphics[width=17cm]{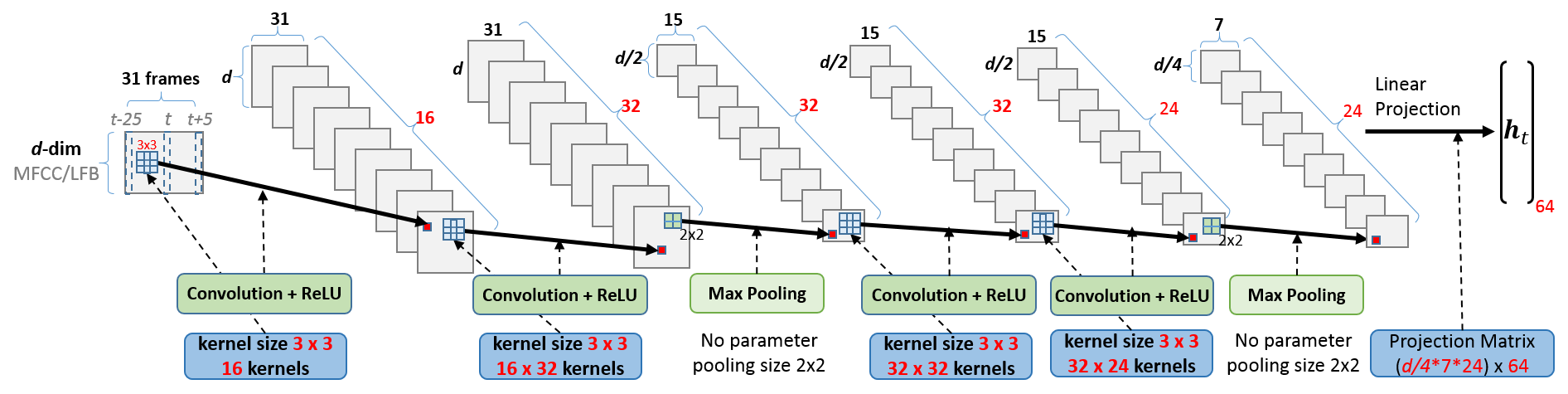}} \vspace{-0.35cm}
\caption{The architecture of the CNN model for exacting the discriminate speaker representations. The green blocks indicate the operations in the network, such as time-frequency convolutions, ReLU activation function, max pooling and full-connected linear projection. The blue blocks show the number of parameters in each operation. The batch normalization is applied. \vspace{-0.35cm}}
\label{fig:cnn}
\end{figure*}

\vspace{-0.1cm}
\subsection{Learning Phonetic Representation}
\label{ssec:asr_dnn}

Previous research has demonstrated that using the phonetic information from the DNNs yields significant gains for SV \cite{richardson2015unified,Zeinali+2016}. These \emph{phonetic discriminative} DNNs refer to the neural networks that can classify each frame of speech as a specific phoneme/senone. Basically, they are DNNs trained for speech recognition \cite{dahl2012context} or keyword spotting \cite{chen2014small}. 
The gain of this approach is mainly due to the phonetic representation extracted from these DNNs can help to align (or normalize) different speakers into the same phonetic space to compare.

\vspace{-0.1cm}
\subsubsection{Phonetic Bottleneck Features ${\bm b}^{\sf phn}_t$}

The most successful and simple approach of using \emph{phonetic discriminative} DNNs for speaker verification is the so-called bottleneck feature approach \cite{fu2014tandem}. These DNNs are learned in the supervised manner to classify the phoneme/senone labels. Once the network is trained, the bottleneck features $\bm b^{\sf phn}_t$ can be extracted for every frame, from the outputs of the last hidden layer of the DNN before the sigmoid nonlinearity \cite{grezl2007probabilistic}. The bottleneck features can also be incorporated together with the input MFCC features as Tandem features  \cite{zhu2004tandem}. These features are then used to train a back-end classifier like a GMM-UBM \cite{richardson2015unified} or i-vector/PLDA \cite{Zeinali+2016}.

\subsubsection{Phonetic Posterior Features ${\bm \gamma}^{\sf{phn}}_t$}

Another approach that makes use of a \emph{phonetic discriminative} DNN for speaker verification is the phonetic posterior feature approach. The basic idea is to replace the frame alignment posteriors ${\bm \gamma}^{\text{UBM}}_t$ generated by the UBM with the senone posteriors ${\bm \gamma}^{\sf{phn}}_t$ produced by the DNN \cite{lei2014novel}.
These posteriors ${\bm \gamma}^{\sf{phn}}_t$ can be directly used to compute the first-order and second-order statistics for i-vector extraction \cite{lei2014novel}. In this work the phonetic posteriors features ${\bm \gamma}^{\sf{phn}}_t$ is also used to provide context information in the attention model described in Section~\ref{sec:attention}.

\subsection{Learning Speaker Representation}
\label{ssec:spk_dnn}

Alternative to the \emph{phonetic discriminative} DNNs, this section discusses the \emph{speaker discriminative} DNNs, which represents a more natural configuration for speaker verification. A \emph{speaker discriminative} DNN is a neural network trained to discriminate between speakers. This type of neural networks has been successfully trained to extract speaker information, such as speaker articulatory features \cite{ZhangMakMeng07} and d-vectors \cite{variani2014deep}.

\subsubsection{dnn-vectors}

In this approach DNNs are trained to discriminate between speakers at the frame-level. Basically, the model tries to classify each frame as belonging to $1$-of-$N$ speakers, where $N$ is the number of background speakers \cite{variani2014deep}. 
After training each frame of an utterance is forward propagated through the network, and the output of the last hidden layer is used to produce an frame-level speaker representation. All the frame-level features are then  averaged to form an utterance-level speaker supervector called a d-vector. 
The main drawback of d-vectors is that the speaker discrimination is achieved based on a time-scale at which phonetic variability is dominant.

\subsubsection{rnn-vectors}
\label{sssec:rnn-vector} \vspace{-0.1cm}
Previously, a RNN based approach for text-dependent speaker verification that makes use of utterance-level features has been proposed \cite{End2EndGoogle,Bhattacharya+2016}. This approach addresses the main criticism of the d-vector approach, i.e., the frame-level speaker classification. In speaker discriminate RNNs, there is a single label associated with each utterance, instead of one for every frame. The hidden vector (after the activation function) of the RNNs in the last frame, ${\bm h}_T$, is an effective summary of the entire utterance. One draw back of this utterance-level ${\bm h}_T$ is it can only be used for text-dependent speaker verification.

\vspace{-0.2cm}
\subsubsection{cnn-vectors}
\vspace{-0.1cm}

Recently, deep CNNs with small kernels have been shown to achieve a better performance than LSTMs in many speech recognition tasks \cite{sercu2016advances,saon2015ibm}. Inspired by these works, a deep CNN architecture is proposed here to extract the speaker discriminative information. The rnn-vector approach described in Section \ref{sssec:rnn-vector} can only be applied to text-dependent speaker verification, while the proposed CNN is suitable for both text-dependent and text-independent tasks.

The CNNs are neural networks with special structures. Fig.~\ref{fig:cnn} illustrates the proposed deep CNN with the VGG-style architecture \cite{VGGnet}. In this CNN the first layer is called a convolution layer, which consists of a number of feature maps. Each neuron in the convolution layer receives input from a local field (e.g. a $3\times3$ window) representing features of a specific frequency range. These local windows shift across time and frequency. The neurons receive different shifted local field as inputs. Neurons in the convolution layer that belong to the same feature map share the same weights, known as kernels or filters. Thus, the convolution layer yield a convolution of the kernels with the inputs. One specialty of our CNN is an asymmetric context window (30 frames in the history and 5 frames in future) is applied to the inputs to control the latency as shown in Fig.~\ref{fig:cnn}. Note the LSTMs actually use the information from ${\bf O}_{1:t}$ for each frame $t$, which can also be viewed as an asymmetric context window. The zero-padding is used during the convolution. The batch normalization \cite{ioffe2015batch} is applied to improve the training convergence. 

%rectified linear units (ReLU)

\vspace{-0.1cm}
\section{Attention Mechanism}
\label{sec:attention}\vspace{-0.1cm}

As discussed in Section \ref{sec:end2end}, rather than simply averaging the frame-level speaker representation $\bm h_t$ from the CNN to produce an utterance-level feature, we propose to learn the best combination of $\bm h_t$. The basic idea is to use an attention network to learn the best way to combine the frame-level speaker features by utilizing the phonetic context information. 

The approach is motivated by the idea of visual attention in the image captioning problem \cite{vinyals2015show}. In the image captioning, when the model is trying to generate the next word of the caption, that word is usually describing \emph{only a part of the image}. Thus the attention network serves as a selection and combination model. In the context of our problem, when the CNN model is trying to generate a speaker feature vector, that vector is only describing the speaker characteristics in a specific context or a specific phonetic space.  Thus the attention network servers as an alignment and combination model. 

Generally, an attention model is an approach that takes $T$ arguments ${\bm h}_1, \ldots, {\bm h}_T$, and a context $\bm c$. It return a vector ${\bm f}$ which is supposed to be the summary of the $\bm h_{1:T}$, focusing on information corresponding to the context $\bm c$. More formally, it returns a weighted mean of the $\bm h_t$, and the weights are chosen according the relevance of each $\bm h_t$ given the context $c$, as shown in Fig.~\ref{fig:att} (a).

\vspace{-0.1cm}
\subsection{attention network with posterior weights} 
\begin{figure}[t]
  \centering
  \centerline{\includegraphics[width=9cm]{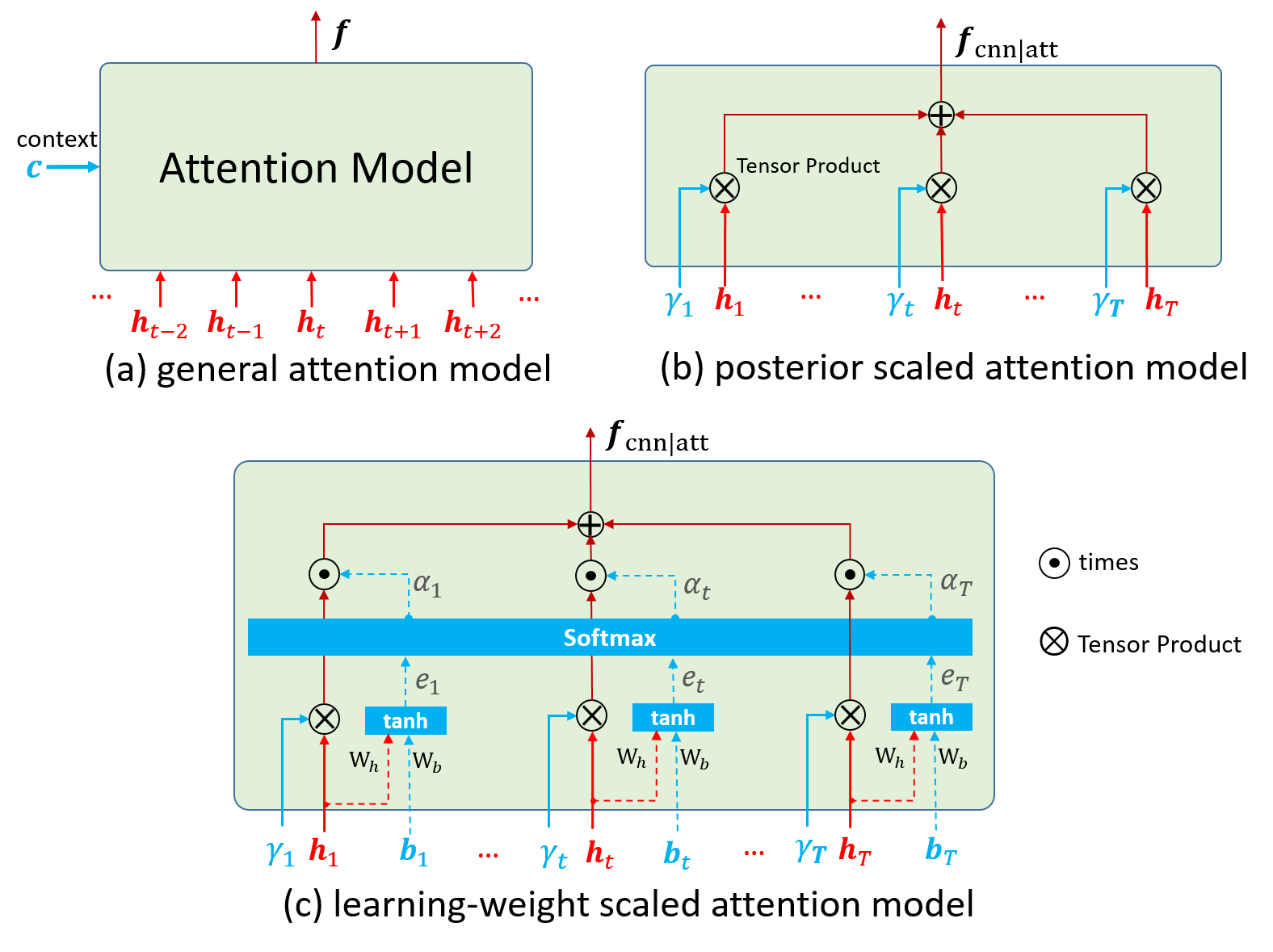}} \vspace{-0.3cm}
\caption{The architecture of the general attention model and two proposed attention networks. The network can learn to align and combine the frame-level speaker representation ${\bm h}_t$ to form a speaker supervector ${\bm f}_{\sf cnn|att}$.} 
\label{fig:att} \vspace{-0.2cm}
\end{figure}

\vspace{-0.1cm}
The first proposed attention network simply uses the DNN posterior features as the context information shown in Fig.~\ref{fig:att} (b). The posterior probabilities from the KWS-DNN naturally provide the alignment information and combination weights. Thus the speaker supervector ${\bm f}_{\sf cnn|att}$ can be computed by
\begin{equation}\label{eq:tensor}
     {\bm f}_{\sf cnn|att} = \sum_{t=1}^T \textcolor{red}{{\bm h}^{\sf spk}_t} \otimes \textcolor{myblue}{{\bm \gamma}^{\sf phn}_t}
    %  = \sum_{t=1}^T {\bm h}_t \otimes 
    %   \begin{bmatrix}
    %     P( {\sf hh} | {\bm o}_t) \\
    %     %P( {\sf ey} | {\bm o}_t) \\
    %     \vdots                   \\
    %     P( {\sf ey} | {\bm o}_t) 
    %   \end{bmatrix}             \\
      = \sum_{t=1}^T 
      \begin{bmatrix}
       \textcolor{myblue}{P( {\sf hh} | {\bm o}_t)} \cdot \textcolor{red}{{\bm h}^{\sf spk}_t}  \\
        %P( {\sf ey} | {\bm o}_t) \\
        \vdots                   \\
       \textcolor{myblue}{P( {\sf ey} | {\bm o}_t)} \cdot \textcolor{red}{{\bm h}^{\sf spk}_t}
      \end{bmatrix}_{640} 
\end{equation}
where $\otimes$ is the Kronecker product \cite{graham1982kronecker}, $\bm h^{\sf spk}_t$ is a 64-dim feature extracted from the speaker discriminative CNN, and feature ${\bm \gamma}^{\sf phn}_t$ has 10 dimensions, each represents a probability of phoneme in 
``Hey Cortana"\footnote{The Hey-Cortana DNN has nine phonemes \{$\sf hh, ey, k, ao, r, t, aa, n, er$\} and one garbage state used to absorb the silence, noise and other phonemes.}, \vspace{-0.3cm}
\[
\textcolor{red}{{\bm h}^{\sf spk}_t} = 
       \bigg[
         {\sf cnn}( {\bm o}_t) 
       \bigg]_{64}
       ,~~
\textcolor{myblue}{{\bm \gamma}^{\sf phn}_t} = 
       \begin{bmatrix}
         P( {\sf hh} | {\bm o}_t) \\
         %P( {\sf ey} | {\bm o}_t) \\
         \vdots                   \\
         P( {\sf ey} | {\bm o}_t) 
       \end{bmatrix}_{10}. \vspace{-0.1cm}
\]
Note that the ${\bm \gamma}^{\sf phn}_t$ is normally a sparse vector. The Eq.\ref{eq:tensor} is basically stacking the speaker vector $\bm h^{\sf spk}_t$ to the corresponding phonetic space, with a probability weight. For example, if $P( {\sf hh} | {\bm o}_1)=1$, then the supervector ${\bm f}_{\sf cnn|att}$ will only have the top $64$-dim non-zero features. By this way, different speaker vectors will be compared only in the same phonetic space. In Eq.\ref{eq:tensor} the frame-level speaker features are weighted by phone posteriors. However the frame combination weights can also be learned through an attention network. This approach will be discussed in the next section.

\subsection{attention network with learned weights}\label{ssec:att2}

In this section, an advanced attention network is proposed. It not only utilizes the alignment information but can also learn the combination weights. This is because different frames may contain different amount of speaker discriminative information. The architecture of this attention network is shown in Fig.~\ref{fig:att}~(c). In this model the context information (shown as blue arrows) contains  two sources, the posterior features ${\bm \gamma}^{\sf phn}_t$ and the bottleneck feature ${\bm b}^{\sf phn}_t$. The sparse vector ${\bm \gamma}^{\sf phn}_t$ is used to mapping the speaker characteristics into the corresponding phonetic space. The bottleneck features ${\bm b}^{\sf phn}_t$ and $\bm h^{\sf spk}_t$ are used together to learn the combination weights for better speaker discrimination. 
\begin{align}
    e_t &= \tanh ( {\bm W}_h \textcolor{red}{{\bm h}^{\sf spk}_t} + {\bm W}_b \textcolor{myblue}{{\bm b}^{\sf phn}_t} ) \\    
    \textcolor{olive}{\alpha_t} &= \frac{\exp{e_t}}{\sum_{j=1}^T \exp{e_j} } \\
    {\bm f}_{\sf cnn|att} & = \sum\limits_{t=1}^T \left( 
                \textcolor{olive}{\alpha_t} \cdot \textcolor{red}{{\bm h}^{\sf spk}_t} \otimes \textcolor{myblue}{{\bm \gamma}^{\sf phn}_t}
                \right)  
\end{align}
where the $ \otimes {\bm \gamma}^{\sf phn}_t$ serves as an alignment model as discribed in previous section and the $e_t$ serves as a combination model. $\alpha_t$ is the softmax distribution over the input sequence. The combination weights represents the density of speaker information in each frame according to the context. The tensor product and the softmax functions are differentiable, therefore the entire network, including the CNN, attention model, logistic regression can be jointly trained using the stochastic gradient descent.

In this work we applied the attention model to the text-dependent speaker verification. However, as the attention model can learn to map the speaker features into the corresponding phonetic space, the proposed approach would work even better for text-independent task.

\section{Experiments and Results}
\label{sec:Exp}

In this section we present the details of network architectures, network training and experimental results. 
All the models investigated in this work were implemented and trained using Theano \cite{2016arXiv160502688short} and the Keras package \cite{chollet2015}.

%The proposed the system is suitable for both text-dependent and text-independent speaker verification, 

\vspace{-0.2cm}
\subsection{Data Sets}
\label{ssec:data}
The proposed approach is evaluated on the Microsoft internal text dependent speaker verification task. A set of utterances beginning with the voice activation phrase ``Hey Cortana" is collected from the Windows 10 desktop Cortana service logs. The “Hey Cortana” segments are taken out using a keyword detector. The length of these segments is around 65 frames to 110 frames. The evaluation comprises about 60k utterances from 3k target speakers and 3k impostors. The enrollment set comprises 6 utterances of “Hey Cortana”. We then selected about 200k utterances from 10k speakers, each with $10$ to $30$ utterances for UBM training. There is no overlap of speakers between the training and testing (enrollment/evaluation) data.

\vspace{-0.2cm}
\subsection{Experiment Setups}
\label{ssec:Exp}

The spectral features used in KWS-DNN consist of $38$ dimensional mel-frequency cepstral coefficients (MFCC) with $13$ filterbanks. The MFCCs contain $12$ static (without energy $C0$), $13$ delta and $13$ delta-delta features. The baseline GMM-UBM and i-Vector systems are using the same MFCCs as KWS-DNNs. A rolling-window based cepstral mean normalization (CMN) and feature warping \cite{pelecanos2001feature} are applied in these systems. The size of the rolling windows is $41$. The CNN model has three channel inputs. The first channel contains $12$-dim static features ($d=12$ in Fig.~\ref{fig:cnn}). The second and third channels use the delta and delta-delta features \cite{abdel2014convolutional}, each has 12 dimensions. To control the run-time latency, an asymmetric context window, $\bm o_{t-25},\dots,\bm o_{t+5}$, is used (see Fig.~\ref{fig:cnn}). 

The KWS-DNN consists of two hidden layers. Each layer has $128$ and $64$ nodes respectively. The output layer has $10$ classes. The GMM-UBM system has $128$ Gaussian mixtures. The i-vector/PLDA system uses $300$ dimensional i-vectors and these are then projected down to $64$ dimensions using the LDA \cite{dehak2011front}. The bottleneck features $\bm b^{\sf phn}_t$ from the KWS-DNN have $64$ dimensions. These bottleneck features are decorrelated using PCA for GMM-UBM systems. The PCA also reduces the dimension of $\bm b^{\sf phn}_t$ to $32$. The CNN model in the end-to-end system uses zero padding and 1-step striding during the convolution operation \cite{sercu2016advances}. A $2 \times 2$ pooling window and 2-step striding are used during the max pooling operation. $N=6$ utterances are selected during the end-to-end network training (see Alg.~\ref{alg:train}). 
This is stimulating the real enrollment scenario as Windows 10 will ask users to enroll 6 utterances in order to allow Cortana try to responds only to the user.
To avoid mismatch between the negative/positive samples ratio in training and evaluation, $T_1=1$ and $T_2=5$ are used in the experiments.
 
\subsection{Results} \vspace{-0.2cm}
\label{ssec:subsubhead}

First, we compare the results between different systems. The equal error rates (EERs) are shown in  Table~\ref{tab:results_allsystem}. The KWS-DNN bottleneck feature is effective for all the systems. The end-to-end system outperforms the GMM and i-vector system by $25\%$ and $9\%$, respectively. Second, we compare different attention mechanisms and speaker discriminate neural networks (see Table~\ref{tab:results_E2E}). The proposed attention network with learned weights (Section \ref{ssec:att2}) works the best. The proposed CNN model also outperforms the DNN (with the same mount of parameters) by $6\%$. Comparing with LSTMs the gain of proposed CNN is very small. This may due to the task is text-dependent speaker verification. Larger gains can be expected for text-independent tasks (future work). Third, we show the importance of using speaker vector pool to select the most similar impostor during the end-to-end training in Table~\ref{tab:results_svpool}. 

%\vspace{-0.2cm}
\begin{table}[h]
\centering
\renewcommand{\tabcolsep}{0.1cm}
\begin{tabular}{l|c|c|c}
\toprule
 Features & GMM-UBM & i-vector/PLDA & End-to-End \\ \midrule
 MFCC $\bm o_t$                    & 5.8\% & 4.7\% & 4.2\%   \\ 
 MFCC $\bm o_t$+BN $\bm b^{\sf phn}_t$       & 4.8\% & 4.3\% & 4.1\%  \\ \bottomrule
\end{tabular}
\caption{The results of the GMM-UBM, i-vector/PLDA and End-to-End systems in EER\%. The CNN model shown in Fig.~\ref{fig:cnn} and the learning-weight scaled attention network shown in Fig.~\ref{fig:att}(c) are used in End-to-End system. 
%BN is short for the KWS-DNN bottleneck features.
}\label{tab:results_allsystem} \vspace{-0.2cm}
\end{table}

%\vspace{-0.2cm}
\begin{table}[h]
\centering
\begin{tabular}{lccc}
\toprule
            & \multicolumn{3}{c}{Speaker Modeling}\\ \cmidrule(r){2-4}
 Attention Mechanism & CNN & DNN & LSTM\\ \midrule
 posterior scaled attention         & 4.5\% & 5.0\% & 4.5\%  \\ 
 learning-weight scaled attention   & 4.1\% & 4.5\% & 4.2\% \\ \bottomrule
\end{tabular}
\caption{EERs of different attention networks and speaker models in End-to-End system with the same $[\bm o_t + \bm b_t]$ features. }\label{tab:results_E2E}\vspace{-0.2cm}
\end{table}

%\vspace{-0.2cm}
\begin{table}[h]
\centering
\renewcommand{\tabcolsep}{0.1cm}
\begin{tabular}{lc}
\toprule
 Training Algorithm  & best End-to-End system \\ \midrule
 random sample test speakers                   &  4.6\%   \\ 
 using speaker vector pool         &  4.1\%  \\ \bottomrule
\end{tabular}
\caption{The effectiveness of speaker vector pool in Alg.~\ref{alg:train}.}\label{tab:results_svpool}\vspace{-0.2cm}
\end{table}

\vspace{-0.3cm}
\section{Conclusion} \vspace{-0.2cm}
\label{sec:conclusion}
A novel end-to-end system for text-dependent SV is described in this paper. 
A speaker discriminative CNN is used to extract frame-level features. Another contribution of this work is the extracted frame-level speaker features are combined through an attention network to generate an utterance-level speaker representation. The proposed attention model takes the speaker discriminative information and phonetic information to learn the attention weights. The third contribution is the whole system, including the CNN and attention models, are joint learned using an end-to-end training algorithm. The algorithm can automatically select the most similar impostors for each target speaker during the training. 
We show the effectiveness of the proposed system on Windows $10$ ``Hey Cortana" speaker verification task.

%\vfill
%\pagebreak

\bibliographystyle{IEEEbib}
\bibliography{strings,refs}

\begin{thebibliography}{10}

\bibitem{Campbell97}
J.~P. {Campbell Jr.},
\newblock ``Speaker recognition: {A} tutorial,''
\newblock {\em Proc. IEEE}, vol. 85, no. 9, pp. 1437--1462, 1997.

\bibitem{Auckenthaler99}
R.~Auckenthaler, E.~Parris, and M.~Carey,
\newblock ``Improving a {GMM} speaker verification system by phonetic
  weighting,''
\newblock in {\em Proc. ICASSP}, 1999, pp. 1440--1444.

\bibitem{Matsui&Furui91}
T.~Matsui and S.~Furui,
\newblock ``A text-independent speaker recognition method robust against
  utterance variations,''
\newblock in {\em Proc. ICASSP}, 1991, pp. 377--380.

\bibitem{RSR2015}
Bin Ma Haizhou~Li Anthony~Larcher, Kong-Aik~Lee,
\newblock ``Text-dependent speaker verification: Classifiers, databases and
  rsr2015.,''
\newblock {\em Speech Communication}, vol. 60, pp. 56--77, 2014.

\bibitem{End2EndGoogle}
Georg Heigold, Ignacio Moreno, Samy Bengio, and Noam~M. Shazeer,
\newblock ``End-to-end text-dependent speaker verification,''
\newblock in {\em International Conference on Acoustics, Speech and Signal
  Processing (ICASSP)}, 2016.

\bibitem{variani2014deep}
Ehsan Variani, Xin Lei, Erik McDermott, Ignacio~Lopez Moreno, and Javier
  Gonzalez-Dominguez,
\newblock ``Deep neural networks for small footprint text-dependent speaker
  verification,''
\newblock in {\em ICASSP}. IEEE, 2014, pp. 4052--4056.

\bibitem{Reynolds95b}
D.~A. Reynolds,
\newblock ``Speaker identification and verification using {G}aussian mixture
  speaker models,''
\newblock {\em Speech Communications}, vol. 17, pp. 91--108, 1995.

\bibitem{Kenny07}
P.~Kenny, P.~Boulianne, G. amd~Ouellet, and P.~Dumouchel,
\newblock ``Joint factor analysis versus eigenchannels in speaker
  recognition,''
\newblock {\em IEEE Transactions on Audio,Speech and Language Processing}, vol.
  15, no. 4, pp. 1435--1447, May 2007.

\bibitem{novoselov2014text}
Sergey Novoselov, Timur Pekhovsky, Andrey Shulipa, and Alexey Sholokhov,
\newblock ``Text-dependent gmm-jfa system for password based speaker
  verification,''
\newblock in {\em ICASSP}. IEEE, 2014, pp. 729--737.

\bibitem{zhang2011optimized}
Shi-Xiong Zhang and Man-Wai Mak,
\newblock ``Optimized discriminative kernel for {SVM} scoring and its
  application to speaker verification,''
\newblock {\em IEEE Transactions on Neural Networks}, vol. 22, no. 2, pp.
  173--185, 2011.

\bibitem{stafylakis2013text}
T~Stafylakis, Patrick Kenny, P~Ouellet, J~Perez, M~Kockmann, and Pierre
  Dumouchel,
\newblock ``Text-dependent speaker recognition using plda with uncertainty
  propagation,''
\newblock {\em matrix}, vol. 500, pp. 1, 2013.

\bibitem{lei2014novel}
Yun Lei, Nicolas Scheffer, Luciana Ferrer, and Mitchell McLaren,
\newblock ``A novel scheme for speaker recognition using a phonetically-aware
  deep neural network,''
\newblock in {\em ICASSP}. IEEE, 2014, pp. 1695--1699.

\bibitem{Bhattacharya+2016}
Gautam Bhattacharya, Patrick Kenny, Jahangir Alam, and Themos Stafylakis,
\newblock ``Deep neural network based text-dependent speaker verification :
  Preliminary results,''
\newblock in {\em Odyssey 2016: The Speaker and Language Recognition Workshop},
  Bilbao, Spain, June 21-24 2016, pp. 9--15.

\bibitem{richardson2015deep}
Fred Richardson, Douglas Reynolds, and Najim Dehak,
\newblock ``Deep neural network approaches to speaker and language
  recognition,''
\newblock {\em IEEE Signal Processing Letters}, vol. 22, no. 10, pp.
  1671--1675, 2015.

\bibitem{Zeinali+2016}
Hossein Zeinali, Lukas Burget, Hossein Sameti, Ondrej Glembek, and Oldrich
  Plchot,
\newblock ``Deep neural networks and hidden markov models in i-vector-based
  text-dependent speaker verification,''
\newblock in {\em Odyssey 2016: The Speaker and Language Recognition Workshop},
  Bilbao, Spain, June 21-24 2016, pp. 24--30.

\bibitem{dehak2010cosine}
Najim Dehak, Reda Dehak, James~R Glass, Douglas~A Reynolds, and Patrick Kenny,
\newblock ``Cosine similarity scoring without score normalization
  techniques.,''
\newblock in {\em Odyssey}, 2010, p.~15.

\bibitem{kenny2013plda}
Patrick Kenny, Themos Stafylakis, Pierre Ouellet, Md~Jahangir Alam, and Pierre
  Dumouchel,
\newblock ``Plda for speaker verification with utterances of arbitrary
  duration,''
\newblock in {\em 2013 IEEE International Conference on Acoustics, Speech and
  Signal Processing}. IEEE, 2013, pp. 7649--7653.

\bibitem{sercu2016advances}
Tom Sercu and Vaibhava Goel,
\newblock ``Advances in very deep convolutional neural networks for {LVCSR},''
\newblock {\em arXiv preprint arXiv:1604.01792}, 2016.

\bibitem{saon2015ibm}
George Saon, Hong-Kwang~J Kuo, Steven Rennie, and Michael Picheny,
\newblock ``The {IBM} 2015 english conversational telephone speech recognition
  system,''
\newblock {\em arXiv preprint arXiv:1505.05899}, 2015.

\bibitem{vaidya1989ano}
Pravin~M Vaidya,
\newblock ``An {O}(nlogn) algorithm for the all-nearest-neighbors problem,''
\newblock {\em Discrete \& Computational Geometry}, vol. 4, no. 2, pp.
  101--115, 1989.

\bibitem{dehak2011front}
Najim Dehak, Patrick~J Kenny, R{\'e}da Dehak, Pierre Dumouchel, and Pierre
  Ouellet,
\newblock ``Front-end factor analysis for speaker verification,''
\newblock {\em IEEE Transactions on Audio, Speech, and Language Processing},
  vol. 19, no. 4, pp. 788--798, 2011.

\bibitem{richardson2015unified}
Fred Richardson, Douglas Reynolds, and Najim Dehak,
\newblock ``A unified deep neural network for speaker and language
  recognition,''
\newblock {\em arXiv preprint arXiv:1504.00923}, 2015.

\bibitem{dahl2012context}
George~E Dahl, Dong Yu, Li~Deng, and Alex Acero,
\newblock ``Context-dependent pre-trained deep neural networks for
  large-vocabulary speech recognition,''
\newblock {\em IEEE Transactions on Audio, Speech, and Language Processing},
  vol. 20, no. 1, pp. 30--42, 2012.

\bibitem{chen2014small}
Guoguo Chen, Carolina Parada, and Georg Heigold,
\newblock ``Small-footprint keyword spotting using deep neural networks,''
\newblock in {\em 2014 IEEE International Conference on Acoustics, Speech and
  Signal Processing (ICASSP)}. IEEE, 2014, pp. 4087--4091.

\bibitem{fu2014tandem}
Tianfan Fu, Yanmin Qian, Yuan Liu, and Kai Yu,
\newblock ``Tandem deep features for text-dependent speaker verification.,''
\newblock in {\em INTERSPEECH}, 2014, pp. 1327--1331.

\bibitem{grezl2007probabilistic}
Frantisek Gr{\'e}zl, Martin Karafi{\'a}t, Stanislav Kont{\'a}r, and Jan
  Cernocky,
\newblock ``Probabilistic and bottle-neck features for lvcsr of meetings,''
\newblock in {\em 2007 IEEE International Conference on Acoustics, Speech and
  Signal Processing-ICASSP'07}. IEEE, 2007, vol.~4, pp. IV--757.

\bibitem{zhu2004tandem}
Qifeng Zhu, Barry Chen, Nelson Morgan, and Andreas Stolcke,
\newblock ``Tandem connectionist feature extraction for conversational speech
  recognition,''
\newblock in {\em International Workshop on Machine Learning for Multimodal
  Interaction}. Springer, 2004, pp. 223--231.

\bibitem{ZhangMakMeng07}
S.~X. Zhang, M.~W. Mak, and H.~M. Meng,
\newblock ``Speaker verification via high-level feature based phonetic-class
  pronunciation modeling,''
\newblock {\em IEEE Trans. on Computers}, vol. 56, no. 9, pp. 1189--1198, 2007.

\bibitem{VGGnet}
K.~Simonyan and A.~Zisserman,
\newblock ``Very deep convolutional networks for large-scale image
  recognition,''
\newblock in {\em ICLR}, 2015.

\bibitem{ioffe2015batch}
Sergey Ioffe and Christian Szegedy,
\newblock ``Batch normalization: Accelerating deep network training by reducing
  internal covariate shift,''
\newblock {\em arXiv preprint arXiv:1502.03167}, 2015.

\bibitem{vinyals2015show}
Oriol Vinyals, Alexander Toshev, Samy Bengio, and Dumitru Erhan,
\newblock ``Show and tell: A neural image caption generator,''
\newblock in {\em Proceedings of the IEEE Conference on Computer Vision and
  Pattern Recognition}, 2015, pp. 3156--3164.

\bibitem{graham1982kronecker}
Alexander Graham,
\newblock ``Kronecker products and matrix calculus: With applications.,''
\newblock {\em JOHN WILEY \& SONS, INC., 605 THIRD AVE., NEW YORK, NY 10158,
  1982, 130}, 1982.

\bibitem{2016arXiv160502688short}
{Theano Development Team},
\newblock ``{Theano: A {Python} framework for fast computation of mathematical
  expressions},''
\newblock {\em arXiv e-prints}, vol. abs/1605.02688, May 2016.

\bibitem{chollet2015}
Fran\c{c}ois Chollet,
\newblock ``Keras,'' \url{https://github.com/fchollet/keras}, 2015.

\bibitem{pelecanos2001feature}
Jason Pelecanos and Sridha Sridharan,
\newblock ``Feature warping for robust speaker verification,''
\newblock in {\em Interspeech}, 2001.

\bibitem{abdel2014convolutional}
Ossama Abdel-Hamid, Abdel-Rahman Mohamed, Hui Jiang, Li~Deng, Gerald Penn, and
  Dong Yu,
\newblock ``Convolutional neural networks for speech recognition,''
\newblock {\em IEEE/ACM Transactions on audio, speech, and language
  processing}, vol. 22, no. 10, pp. 1533--1545, 2014.

\end{thebibliography}

\end{document}